# Dynamic Realms: 4D Content Analysis, Recovery and Generation with Geometric, Topological and Physical Priors

ZHIYANG DOU, The University of Hong Kong, Hong Kong SAR; University of Pennsylvania, U.S.A.

## 1 INTRODUCTION & RESEARCH PROGRESS

My research focuses on the analysis, recovery, and generation of 4D content, where 4D encompasses three spatial dimensions ($x, y, z$) and a temporal dimension $t$, such as **shape** and **motion**. This focus extends beyond static objects to encompass dynamic changes over time, providing a comprehensive understanding of both spatial and temporal variations. These techniques are critical in applications such as AR/VR, embodied AI, and robotics. My research aims to make **4D content generation** more efficient, accessible, and of higher quality by injecting *geometric*, *topological*, and *physical* priors. I also aim to develop effective methods for 4D content **recovery** and **analysis** with these priors.

### 1.1 3D Shape Analysis, Recovery, and Generation

**Shape Analysis.**

♦ *Skeletonize Shapes.* In shape analysis, the medial axis transformation (MAT) is considered a powerful geometric tool for shape abstraction [3], recognition, and generation. In my earlier research, I proposed **Coverage Axis** [1][1], a new

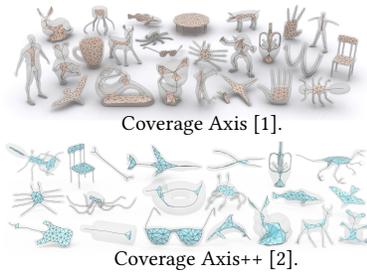
Coverage Axis [1].

Coverage Axis++ [2].

*formulation* that defines a compact and expressive approximation of the MAT for a given three-dimensional shape based on the Set Cover Problem. As shown in Fig. 1, this formulation allows for the global consideration of the overall shape structure, resulting in efficient high-level abstraction and strong noise robustness. It handles general inputs, such as point clouds or low-quality meshes. Recently, we developed an approach from a statistical point-of-view to achieve a compact representation of 3D objects named **Coverage Axis++** [2], which enables skeletonization for various shape inputs, specification of skeletal point numbers, few hyperparameters, and highly efficient computation with improved reconstruction accuracy.

♦ *Orientate Complex Shapes.* Point cloud representations are increasingly popular with the increasing accessibility of point cloud data. Estimating normals with globally consistent orientations for raw point clouds is useful in many downstream tasks, such as segmentation and reconstruction. Despite significant efforts in the past decades, handling unoriented point clouds with various imperfections poses significant challenges, especially when data sparsity is combined with neighboring gaps or thin-walled structures.

We introduce Globally Consistent Normal Orientation (**GCNO**) [4][2] for orienting point clouds. Our approach involves developing a smooth objective

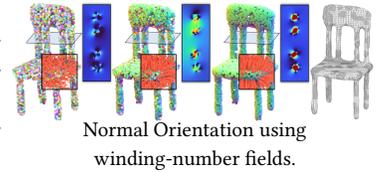
Normal Orientation using winding-number fields.

function to define the requirements of an acceptable Winding-Number Field, which facilitates the identification of globally consistent normal orientations, even in the presence of noisy normals. GCNO significantly outperforms previous methods, especially in handling sparse and noisy point clouds, as well as shapes with complex geometry and topology.

**Shape from X.**

♦ *Shape from Monocular Images.* Acquiring high-quality 3D shapes from monocular images is a challenging yet crucial task for analysis and generation. To tackle this problem, we created Won-

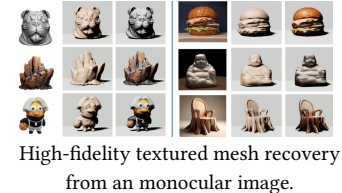
High-fidelity textured mesh recovery from an monocular image.

der3D [5], which offers a cross-domain diffusion model. Wonder3D generates consistent multi-view normal and color maps with a geometry-aware normal fusion algorithm to extract high-quality surfaces from the multi-view 2D representations. Wonder3D greatly outperforms previous works in terms of reconstruction quality, generalization robustness, and efficiency.

♦ *Shape from Point Clouds.* In addition to monocular images, point clouds serve as another important modality for shape acquisition. In our work Reconstructing Feature-line Equipped Polygonal Surface (RFEPS) [6], we use Optimal Transportation to identify edge zones and produce line-type geometric features. Sharp edges and corners are captured, contributing to the feature lines. A Restricted Power Diagram is employed to interpolate the augmented points, prioritizing connections between edge points, leading to effective feature line-equipped surface reconstruction.

**Shape Generation.**

♦ *Shapes with Arbitrary Topologies.* The interplay between topology and geometry generates diverse 3D shapes. However, current approaches often struggle with limited topological and geometric complexity. We introduced **Surf-D** [7], a novel method for generating high-quality shapes as surfaces with arbitrary topologies using diffusion models. We chose the Unsigned Distance Field (UDF) as the representation, as it excels at handling arbitrary topologies and

---
[1]Top cited article in CGF during 2022-2023.

Author's address: Zhiyang Dou, The University of Hong Kong, Hong Kong SAR; and University of Pennsylvania, U.S.A., frankzydou@gmail.com.

[2]SIGGRAPH 2023 Best Paper Award

enables the generation of complex structures. We learn a Continuous Distance Field through Point-to-UDF training, which effectively captures complicated geometry. A latent diffusion model is employed to acquire the distribution of various shapes. Surf-D excels in various tasks, such as conditional/unconditional generation and reconstruction.

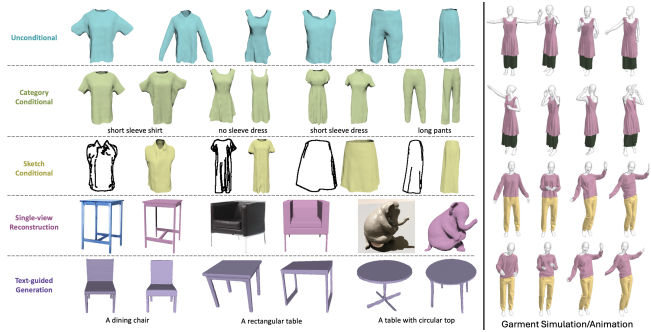

Fig. 1. High-quality surface generation for arbitrary topologies [7].

♦ *Animatable Shapes from Text.* We propose a novel text-to-avatar generation method, SO-SMPL [8], in which we designed a framework based on Score Distillation Sampling to separately generate the human body and garments using explicit geometry offsets to model different garment layers. The generated avatar can be seamlessly integrated into a standard modern CG pipeline for high-quality character animation and garment simulation.

### 1.2 Motion Generation and Analysis

**Motion Generation.** Motion generation is essential for various applications such as the metaverse, embodied AI, human behavior simulation, and crowd analysis. My work has improved the *controllability*, *efficiency*, and *physical plausibility* of generative motion synthesis. Recently, I have focused on collective behavior simulation as well as analysis [9] and quadruped simulation [10].

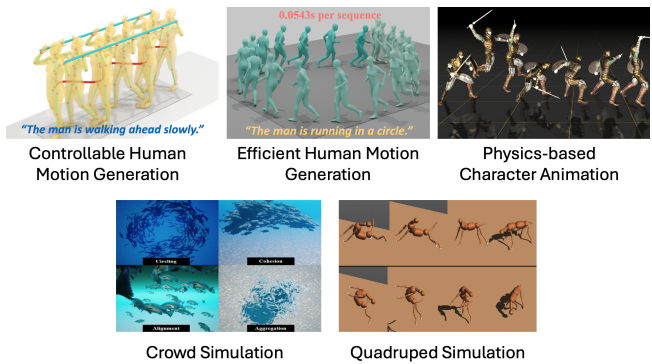

Fig. 2. Our works on controllability [11], efficiency [12] and physical plausibility [13] of human motion generation as well as crowd simulation [9] and quadruped simulation [10].

♦ *Controllability.* Existing methods focus primarily on language or full trajectory control, leaving out precision in synthesizing motions aligned with user-specified trajectories and text descriptions. We created **TLControl** [11], a new method for motion synthesis, incorporating both low-level trajectory and high-level language semantics controls through *an integration of neural-based and optimization-based techniques*. TLControl significantly outperforms SOTAs in terms of trajectory accuracy and time efficiency.

♦ *Efficiency.* State-of-the-art (SOTA) generative diffusion models have achieved impressive motion generation results but struggle to balance speed and quality. We created **EMDM** [12], which captures the complex denoising distribution for a few-step sampling in the diffusion model. This considerably accelerates generation with much fewer sampling steps, achieving a 100x speedup compared to existing SOTA models, enabling high-quality real-time motion generation.

♦ *Physical plausibility.* Our work **C·ASE** [13] presented an efficient and effective framework for learning *Conditional Adversarial Skill Embedding* for physics-based characters. We proposed Focal Skill Sampling and Skeletal Residual Forces to enhance the Generative Adversarial Imitation Learning process. Our simulated character learns a wide range of skills while allowing precise control over which skills are performed.

♦ *Collective Behavior Simulation.* Reproducing realistic collective behaviors is challenging due to limitations in rule-based and existing imitation learning methods. We present Collective Behavior Imitation Learning (**CBIL**) [9], which learns fish schooling directly from videos without needing ground truth trajectories. Our approach uses a Masked Video AutoEncoder for self-supervised representation learning, mapping 2D observations to compact implicit states. We introduce a novel adversarial imitation learning technique to capture complex fish movements, incorporating bio-inspired rewards and priors for stable training. CBIL enables various animation tasks and is effective across different species. We also demonstrate its application in detecting abnormal fish behavior in wild videos.

**Motion Capture&Analysis.** I have been working on human mesh recovery for the body [14], hand, and face [15]. My previous work on efficient human mesh recovery, **TORE** [14], identified token redundancy issues in existing Transformer-based methods (SOTAs) for human mesh recovery. We proposed token reduction strategies by integrating insights from the human body hierarchy and image features into the Transformer design. TORE achieves the SOTA performance with significantly reduced computational costs.

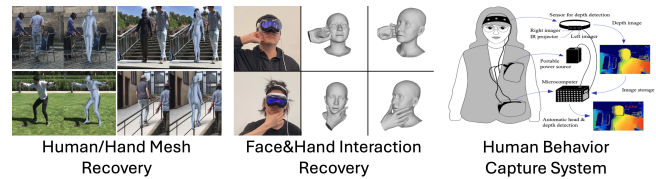

I have been developing hardware devices and machine learning systems to more effectively and efficiently capture human activities, with a focus on close-contact behaviors. During the COVID-19 pandemic, we conducted studies on human close-contact behavior (capture, modeling, and simulation) in various settings, including subways [16], universities [17–19][3], hospitals [20], airports [21], and trains [22]. These efforts contributed to the analysis of virus spread patterns, providing valuable insights for public health.

---

[3]The research [17] was featured in a press release by AAAS EurekAlert!.

## 1.3 Future Directions

My future work will continue to explore advanced techniques for incorporating geometric, topological, and physical priors into 4D content processing and generation:

♦ **Scene-level Physics-aware Shape Generation and Motion Synthesis:** I will develop more effective methods for *integrating geometric, topological, and physical priors into the processes of shape and motion generation*. This includes developing techniques such as Repulsive Shells [Sassen et al.2024] (geometry&physics), Projective Dynamics (physics), and Persistent Homology (topology).

♦ **Capturing and Predicting Complex 4D Dynamics:** While some approaches utilize physical, topological, and geometric priors for 4D modeling, these methods are typically not well-integrated with generative models, particularly in neural-based approaches. My goal is to develop new techniques that can more *efficiently* and *scalably* incorporate these priors during the modeling of complex dynamics, such as deformations, collisions, and interactions between multiple objects.

♦ **4D Content Generation for Robot Learning:** As generated 4D content with integrated priors aids in a robot's learning about the world, I will explore *utilizing generative 4D content (shapes/motions) for robot learning, while reciprocally enhancing the perception and control of the agents during robot-environment interactions*.

## ACKNOWLEDGMENTS

I am deeply grateful to my Ph.D. advisors, Prof. Wenping Wang, Prof. Taku Komura, and Prof. Lingjie Liu, whose knowledge and encouragement have continuously guided me throughout my Ph.D. journey. I am also thankful to my brilliant collaborators, whose insights and dedication have been invaluable during this wonderful experience.